\documentclass[journal]{IEEEtran}

\IEEEoverridecommandlockouts                              

\usepackage[numbers]{natbib}
\usepackage[utf8]{inputenc} 
\usepackage[T1]{fontenc}    
\usepackage{hyperref}       
\usepackage{url}            
\usepackage{booktabs}       
\usepackage{amsfonts}       
\usepackage{nicefrac}       
\usepackage{microtype}      
\usepackage{lipsum}
\usepackage{graphicx}
\usepackage{amsmath}
\DeclareGraphicsExtensions{.png,.jpeg,.pdf}

\title{\LARGE \bf
Towards Brain-inspired System: Deep Recurrent Reinforcement Learning for Simulated Self-driving Agent
}

\author{Jieneng Chen$^{1}$, Jingye Chen$^{2}$, Ruiming Zhang$^{1}$ and Xiaobin Hu$^{3}$ 
\thanks{*This work was financially supported by the German Research Foundation (DFG) and the Technical University of Munich (TUM) in the framework of the Open Access Publishing Program. This research was also funded by the Chinese Ministry of Education's National University Student Innovation and Entrepreneurship Training Program (2018).}
\thanks{$^{1}$Jieneng Chen and Ruiming Zhang are with College of Electronics and Information Engineering, Tongji University, 4800 Cao An Highway, Shanghai, China
        {\tt\small $\{$chenjn, 1631524$\}$@tongji.edu.cn}}%
\thanks{$^{2}$Jingye Chen is with    Department of Computer Science, 
   Fudan University, Shanghai, China 
        {\tt\small 15307130116@fudan.edu.cn}}%
\thanks{$^{2}$Xiaobin Hu is with    Department of Computer Science, 
   Technical University of Munich, Munich, Germany
        {\tt\small xiaobin.hu@tum.de}}%
        
}

\begin{document}
\maketitle
\thispagestyle{empty}
\pagestyle{empty}

\begin{abstract}
An effective way to achieve intelligence is to simulate various intelligent behaviors in the human brain. In recent years, bio-inspired learning methods have emerged, and they are  different from the classical mathematical programming principle. In the perspective of brain inspiration, reinforcement learning has gained additional interest in solving decision-making tasks as increasing neuroscientific research demonstrates that significant links exist between reinforcement learning and specific neural substrates. Because of the tremendous research that focuses on human brains and reinforcement learning, scientists have investigated how robots can autonomously tackle complex tasks in the form of a self-driving agent control in a human-like way. In this study, we propose an end-to-end architecture using novel deep-Q-network architecture in conjunction with a recurrence to resolve the problem in the field of simulated self-driving. The main contribution of this study is that we trained the driving agent using a brain-inspired trial-and-error technique, which was in line with the real world situation. Besides, there are three innovations in the proposed learning network: raw screen outputs are the only information which the driving agent can rely on, a weighted layer that enhances the differences of the lengthy episode, and a modified replay mechanism that overcomes the problem of sparsity and accelerates learning. The proposed network was trained and tested under a third-partied OpenAI Gym environment. After training for several episodes, the resulting driving agent performed advanced behaviors in the given scene. We hope that in the future, the proposed brain-inspired learning system would inspire practicable self-driving control solutions.
\end{abstract}


\section{Introduction}
Recently, research in brain science has gradually received the public's attention. Given the rapid progress in brain imaging technologies and in molecular and cell biology, much progress has been made in understanding the brain at the macroscopic and microscopic levels. Currently, the human brain is  the only truly general intelligent system that can cope with different cognitive functions with extremely low energy consumption. 
Learning from information processing mechanisms of the brain is clearly the key to building stronger and efficient machine intelligence \citep{poo2016china}. In recent years, some bio-inspired intelligent methods have emerged, and they are definitely different from the classical mathematical programming principle. Bio-inspired intelligence has the advantages of strong robustness and efficient, good distributed computing mechanism. Besides, it is easy to combine with other methods.

The mammalian brain has multiple learning subsystems.   Niv et al. \cite{niv2009reinforcement} categorized major learning components into four classes: the neocortex, the hippocampal formation (explicit memory storage system), the cerebellum (adaptive control system) and the basal ganglia (reinforcement learning ). Among these learning components, reinforcement learning is particularly attractive to researches. Nowadays, converging evidence links reinforcement learning to specific neural substrates, thus assigning them to precise computational roles. Most notably, much evidence suggests that the neuromodulator known as dopamine provides basal ganglia target structures with phasic signals that convey a reward prediction error which can influence learning and action selection, particularly in stimulus-driven habitual instrumental behaviors \citep{rivest2005brain}.
Hence, many efforts have been made to investigate the capability of bio-inspired reinforcement learning by applying them to artificial intelligence related tasks. \citep{mcpartland2008learning, peters2008reinforcement, duan2010new, mnih2015human, zhu2017target, gu2017deep}

In recent years, deep reinforcement learning has contributed to many of the spectacular success stories of artificial intelligence in recent years \citep{kober2013reinforcement, henderson2018deep}. After the initial success of the deep Q network (DQN) \citep{mnih2013playing}, a variety of improved models have been published successively. Later on and based on the former discoveries, Mnih et al. \cite{mnih2015human} proposed the Nature DQN in 2015  and introduced the replay memory mechanism to break the strong correlations between the samples. Mnih et al. \cite{mnih2016asynchronous} proposed a deep reinforcement learning approach, in which the parameters of the deep network are updated by multiple asynchronous copies of the agent in the environment. Van Hasselt et al. \cite{van2016deep} suggested the Double DQN to eliminate overestimation; they added a target Q network independent from the the current Q network. It was shown to apply to large-scale function approximation \cite{van2016deep}. Newer techniques included deep deterministic policy gradients and mapping an observation directly to action, both of which could operate over continuous action spaces \cite{lillicrap2015continuous}. Schaul et al. \cite{schaul2015prioritized} suggested prioritized replay, adding prior to the replay memory to relieve the sparse reward and slow converge the problem slowly \citep{schaul2015prioritized}. In the case of partially observable states, the recurrent neural network (RNN) and long short-term memory (LSTM) have been proved to be effective in processing sequence data \citep{hochreiter1997long}. 
\citet{hausknecht2015deep} replaced the last fully connected layer in the network with an LSTM layer. Also,  Foerster et al. \cite{foerster2016learning} proposed to use multi-agent to describe a state distributively. However, most previous studies such as Atari games focused on the simple environment and action space. Because of this limitation, there is an urgent need to further improve the capability of deep reinforcement learning in a more challenging and complex scenario such as simulated 3D driving control problem.

One deep-learning-based studied the simulated self-driving game \citep{honeuralkart}. However,  three problems existed in their implementation. First, they created a handcrafted dataset. Obviously, one can never create this ideal benchmark dataset that includes all the bad situations encountered by the driving agent during training. At best, one can include the best behavior that the driving agent should implement in each step. The driving agent was reported to perform well when it had a good position in the driveway. However, the behavior deteriorated rapidly when the driving agent deviated from the driveway. Such behaviors indicated the dissimilar distribution and instability even though correctional measures were taken on the dataset. Second, they trained and supervised their network in a supervised way. As there are many possible scenarios, manually tackling all possible cases using supervised learning methods will likely yield a more simplistic policy \citep{shalev2016safe}. Third, their experiments were built on ideal conditions; for example, they assumed that the brakes were ignored. In our experiments, we take the brakes into consideration. Moreover, to support autonomous capabilities, a robotic driven agent should adopt human driving negotiation skills when overtaking, halting, braking, taking left and right turns, and pushing ahead in unstructured roadways. It comes naturally that a trial-and-error way of study is more suitable for this simulated self-driving game. Hence, the bio-inspired reinforcement learning method in the study is a more suitable way for the driving agent to learn how to make decisions. 

In our study, we proposed a deep recurrent reinforcement learning network to solve simulated self-driving problems. Rather than creating a handcrafted dataset and training in a supervised way, we adopted a bio-inspired trail-and-error technique for the driving agent to learn how to make decisions. Furthermore, this paper provides three innovations. First, intermediate game parameters were completely abandoned, and the driving agent relied on only raw screen outputs. Second, a weighting layer was introduced in the network architecture to strengthen the intermediate effect. Third, a simple but effective experience recall mechanism was applied to deal with the sparse and lengthy episode.

The rest of this study is organized as follows: Section \ref{sec:methods} describes deep Q-learning, recurrent reinforcement learning, network architecture and implementation details. Section \ref{sec:exp} verifies experimental results. The conclusion of this study is drawn in Section \ref{sec:conclusion} .

\section{Methodology}
 Deep Q-learning is used to help AI agents operate in environments with discrete actions spaces. Based on the knowledge of Deep Q-learning, we proposed a modified DRQN model in order to infer the full state in partially observable environments.

\label{sec:methods}
\subsection{Deep Q-learning}
 Reinforcement learning manages learning policies for an agent interacting in an unknown environment. In each step, an agent observes the current states of the environment, makes decisions according to a policy $\pi$, and observes a reward signal $r_{t}$ \citep{lample2017playing}. Given the current states and a set of available actions, the main aim of the DQN is to approximate the maximum sum of discounted rewards. The value of a given action-state pair or the ‘Q-value’ is given by: 
\begin{equation}
Q^{*}\left(s_{t}, a_{t}\right)=m a x R_{t+1}\label{eq:01}
\end{equation}
According to the Bellman equation, it gives the approximating form of Q-values by combining the reward obtained with the current state-action pair and the highest Q-value at the next state  $S^{'}$, and the best action $a^{'}$:
\begin{equation}
Q(s, a)=Q(s, a)+\gamma * \max _{a^{\prime}} Q\left(s^{\prime}, a^{\prime}\right)\label{eq:02}
\end{equation}
We often use the form involving a iterative process: 
\begin{equation}
Q(s, a)=Q(s, a)+\alpha\left(r+\gamma * \max _{a^{\prime}} Q\left(s^{\prime}, a^{\prime}\right)-Q(s, a)\right)\label{eq:03}
\end{equation}
Actions are typically chosen following a $\varepsilon$-greedy exploration policy. The epsilon value $\varepsilon$ was valued between 0.0 to 1.0. In order to encourage the agent to explore the environment, the $\varepsilon$ was set to 1.0 at the beginning. During the training process, the value decayed gradually as the experience accumulated. Then, the agent could use the experience to achieve the task. 
\begin{equation}
P(a | s)=\frac{\varepsilon}{n(A)}+1-\varepsilon * \operatorname{argmax}_{a} Q(s, a)\label{eq:04}
\end{equation}
The network stands for the approximation of the expected Q-value $Q^*$, which leads to the loss function:
\begin{equation}
\operatorname{Loss}(\theta)=\sum\left(y_{t}-Q_{\theta_{t}}(s, a)\right)^{2}\label{eq:05}
\end{equation}

\subsection{Recurrent Reinforcement Learning}

\begin{figure*}[h!]
  \centering
  \includegraphics[width=1.0\textwidth]{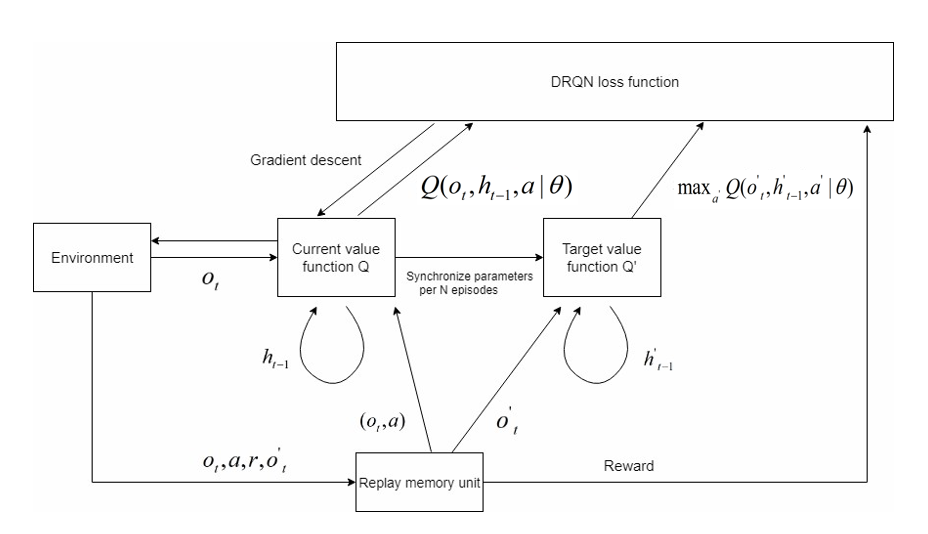}
  \caption{The modified DRQN model. The value function was divided into two categories: the current value function $Q$ and target value function $Q^{'}$. The parameters in $Q$ were assigned to $Q^{'}$ per N episodes. The state contained two elements: $o_{t}$ gained from current environment and $h_{t-1}$ gained from former information. The agent performed action $a$ using a specific policy, and the sequence $ \left(o_{t},a,r,o_{t}^{'}\right) $ was stored in the replay memory unit. We used a prioritized experience replay memory unit here. During training, the sequence was randomly chosen from the replay memory unit. We trained the network using gradient descent to make the current value function $Q$ approach $Q^{'}$ given a specific sequence. The loss function was shown in Eq. \ref{eq:05}.}
  \label{fig:1}
\end{figure*}

\begin{figure*}[h!]
  \centering
  \includegraphics[width=1.0\textwidth]{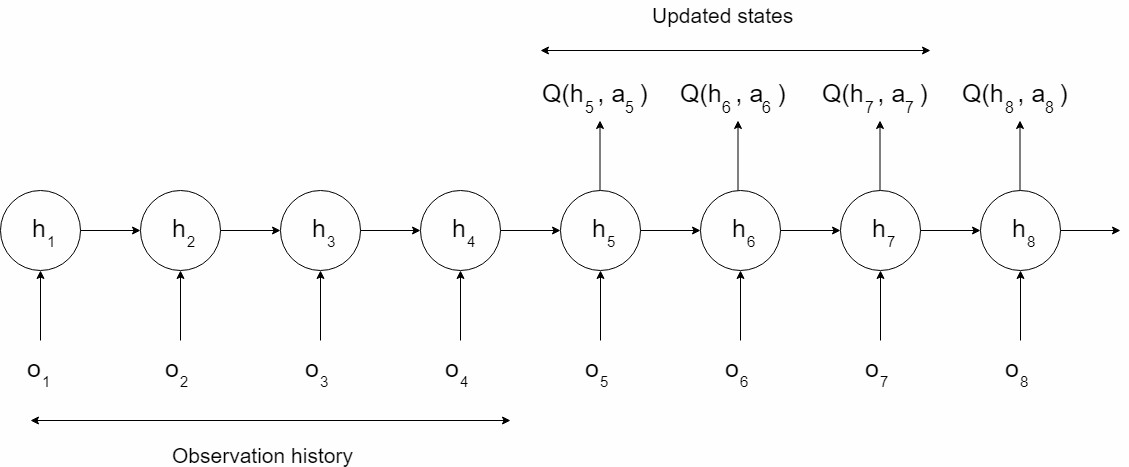}
  \caption{Sequence updates in the recurrent network. Only the scores of the actions taken in states 5, 6 and 7 will be updated. The first four states provide a more accurate hidden state to the LSTM, while the last state provides a target for state 7. }
  \label{fig:2}
\end{figure*}

For some special games which are three-dimensional and partially observable, the DQN lacks the ability to solve the problem. In partially observable environments, the agent only receives an observation $o_{t}$ of the current environment, which is usually insufficient to infer the full state of the system. The real state $s_{t}$ is the combination of the current observation $o_{t}$ and an unfixed length of history states. Hence, we adopted the DRQN model on top of the DQN to deal with such conditions (see Figure \ref{fig:1}). The last full connect layer was replaced by the LSTM in the DRQN model in order to record former information. Figure \ref{fig:2} shows the sequential updates in the recurrent network. An additional input $h_{t-1}$ stood for the previous information is added to the recurrent model. The output of the LSTM $ z\left(o_{t}, h_{t-1}\right) $ , which combined the current observation $o_{t}$ and the history information $h_{t-1}$, was used to approximate the Q-value $ Q\left(o_{t}, h_{t-1}, a_{t}\right) $. The history information was updated and passed through the hidden state to the network in the next time step:
\begin{equation}
h_{t}=L S T M\left(h_{t-1}, o_{t}\right)\label{eq:06}
\end{equation}

\subsection{Network Architecture}
\begin{figure*}[h!]
  \centering
  \includegraphics[width=1.0\textwidth]{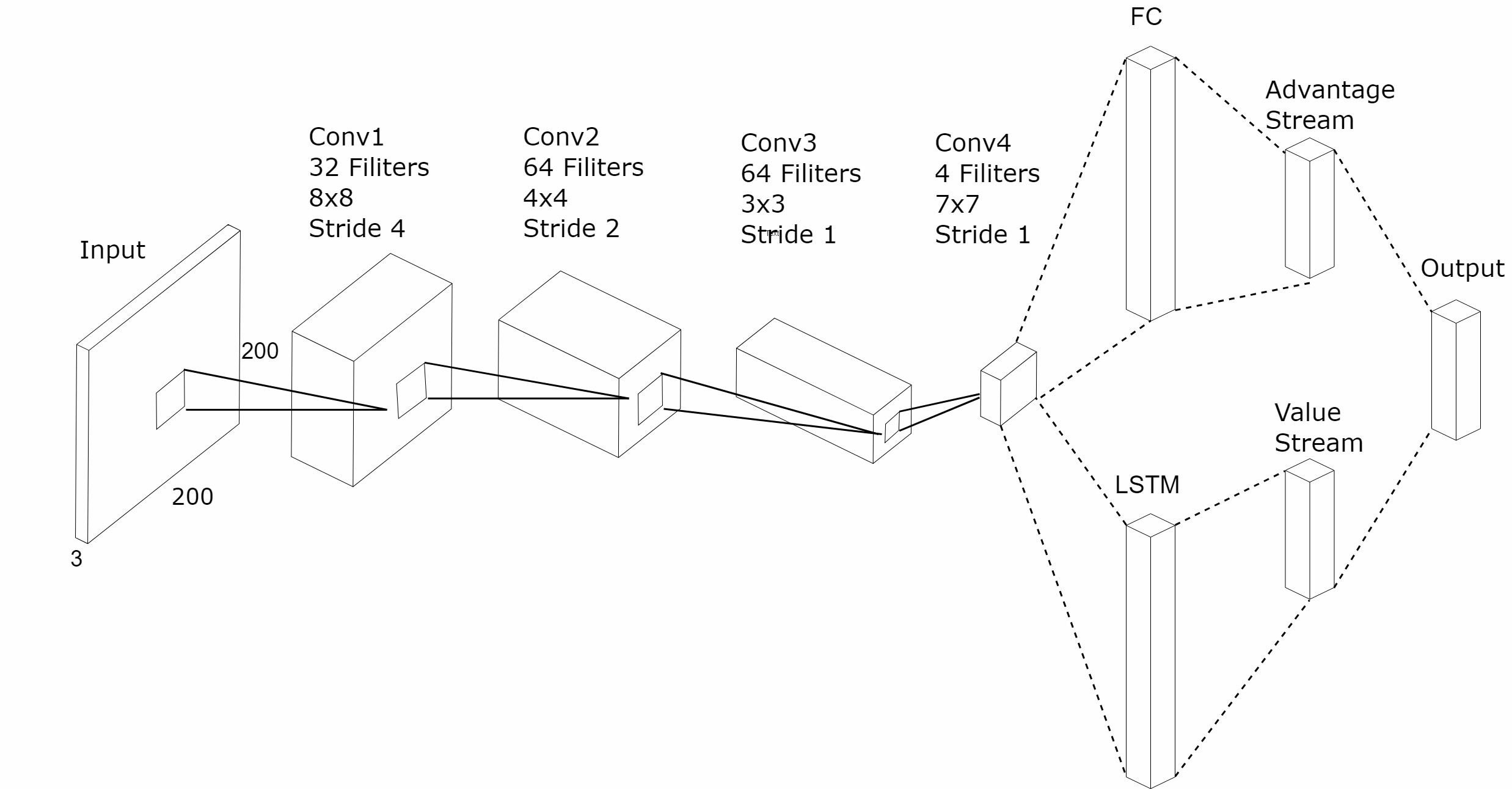}
  \caption{An illustration of the architecture of our model. The input image is assigned to four convolution layers. The output of the convolution layers is split into two streams. The first stream (bottom) flattens the output and feeds it to an LSTM. The second one (top) flattens the output and feeds it to a full connect layer. Then, we obtained an advantage stream and value steam individually and multiplied them as the output. We stored the information in prioritized experience memory unit. As was shown in Figure  \ref{fig:1}, the network was trained using the DRQN loss function. }
  \label{fig:3}
\end{figure*}

In the beginning, we used the baseline DRQN model carried out by the CMU in VizDoom2016 \citep{lample2017playing}. However, we obtained unsatisfied results with the same model. Hence, we have made three improvements in our modified model to make things work better. The whole architecture is shown in Figure \ref{fig:3}. First, the network was built on top of the NVIDIAs autopilot model \citep{krizhevsky2012imagenet}. To reduce overfitting, the original model was modified by adding several batch normalization layers. We used a four-layer stronger CNN for feature extraction. The input size was set to 320*240. The first convolutional layer contained 32 kernels, with the a size of 8*8, and the stride was 4. The second convolutional layer contained 64 kernels, with the a size of 4*4 and the stride was 4. The third layer contained 128 kernels, with the a size of 3*3 and the stride was 1. The last convolutional layer contained 256 kernels with a size of 3*3 and the stride is 1. The activate functions were all Relu, and the size of pooling layers were all 2*2. Second, we abandoned the fully connected layer before the LSTM layer in the original DRQN model and fed the LSTM directly with the high-level feature. The number of units in LSTM was set to 900. Third, the subsequent structure was divided into two groups for different purposes. One  was mapped to the set of possible actions, and the other was a set of scalar values. The final action value function was acquired using both of them. We will introduce their functions respectively.

We used two collateral layers rather than a single DRQN network to approximate the value function. The auxiliary layer also had the same dimension as the action space. The original intention was to balance the impact of history and the current. The DRQN model enabled the driving agent to make a reasonable action from a fully observing perspective. Nevertheless, we wanted to focus our attention on the precise instantaneous change of the current view. The auxiliary layer was mapped to [0, 1.0] using the softmax method, thus suggesting the direction of correction for the raw approximations using DRQN. Because of this intervention, the network would not only learn the best action for a state but also understand the importance of taking actions. Imagine an agent driving in a straight line and having an obstacle far ahead, an original DRQN model could learn that it is time to make the driving agent move a bit to avoid hitting the obstacle. The modified model will also have an insight into when it is the best time to move, with the knowledge that the danger of the obstacle increases as it gets closer. $ V(s, a) $ was used to represent the original output of the DRQN, and $ A(s, a) $ was used to represent the significance provided by the auxiliary layer. We used the formula to express the final strengthen of the Q-value (see Figure \ref{fig:3}):
\begin{equation}
Q(s, a)=V(s, a)^{T} * A(s, a)\label{eq:07}
\end{equation}

The result was stored in the prioritized experience memory unit. During training, the sequence in the memory unit was removed, and we used the Eq. \ref{eq:05} to calculate the loss.

\subsection{Implementation Details}
Reinforcement learning consists of two basic concepts: action and reward. Action is what an agent can do in each state. Given that the screen is the input, a robot can take steps within a certain distance. An agent can take  finite actions. When a robot takes an action in a state, it receives a reward. Here, the term reward is an abstract concept that describes the feedback from the environment. A reward can be positive or negative. When the reward is positive, it corresponds to our normal meaning of reward. However, when the reward is negative, it corresponds to what we usually call punishment. We also describe the training details such as the hyperparameters, input size selection, frameskip, and prioritized replay.

\subsubsection{Action Space}
The game had five actions, including Left, Right, Straight, Brake and Backwards. The rocking range of the joystick reflects the numerical value of the speed control, which is mapped into a region of -80 to 80. To keep the model simple, we considered that only the turning control had a high requirement for precision. The speed section was discretized into a set of [0, 20, 40, 80] for speed control. Another three actions were represented by 1/0 flag. Thus, each action was represented by a 5-dimensional vector shown below:

$$ 
\begin{aligned} \text {$actions$} &=[-40,0,1,0,0], \text { $left$ } \\ &=[40,0,1,0,0], \text { $right$ } \\ &=[0,0,0,1,0], \text { $go$ $backwards$ } \\ &=[0,0,1,0,0], \text { $go$ $straight$ } \\ &=[0,0,0,0,1], \text { $brake$ } \end{aligned}
$$

\subsubsection{Reward}
Under most circumstances, the driving agent cannot explore a path with big rewards initially. The driving agent often gets stuck somewhere in the halfway and wait for the time to elapse before resetting. We have to make the rewards of these cases variant in order to make these experience meaningful. Hence, we set a series of landmarks along the track. A periodical reward was given to the driving agent when each landmark was reached. The closer the distance between the landmark and the destination, the bigger the phased reward was given. We have considered formulating a more precise reward system to increase density, such as giving the driving agent a slight punishment when it deviates from the driveway. However, hidden game parameters were needed, and they were against the original aim of this research.

\subsubsection{Hyperparameters}
The network was trained using the RMSProp algorithm and the minibatch size was 40. The replay memory was set to contain 10000 recent frames. The learning rate $\alpha$ was set, from 1.0 to 0.1, to follow a linear degradation and then fixed at 0.1. The discount factor $\gamma$ was set to 0.9. The exploration rate $\varepsilon$ for the $\varepsilon$-greedy policy was set to follow a linear degradation, from 1.0 to 0.1, and then fixed at 0.1. The exploration rate $\varepsilon$ was set to 0 when we evaluated the model. The original screen outputs were three channel RGB images. They were first transformed into grey scale images and then fed to the network.
\subsubsection{Input size}
In the beginning, the original 640*480 screen resolution was resized to 160*120 to accelerate training. After several hours of trials, the driving agent still got stuck in most cases and could not complete one lap. The resulting rewards oscillated around the baseline reward for not finishing the game in the limited number of steps, thus indicating the resolution was too low to recognize.
Then, the input size was resized to 320*240. We also reduced the punishments to encourage positive rewards. After the observation of the same length of time, the distribution of the resulting rewards cast off random oscillation and started to turn positive over the baseline. Hence, we kept this resolution as the input size of the network despite the memory consumption as the system started to learn within the acceptable limit of time.

\subsubsection{Frameskip}
Obviously, it is not necessary to feed every single frame into the network because there is only a weak change between adjacent frames. Because of this, we used the frame-skip technique \citep{bellemare2013arcade} as in most approaches. One frame was taken out as the network input by every k + 1 frames. The same action was repeated over the skipped frames. The higher k was set, the faster the training speed. However, the control effect became imprecise at the same time. Considering the balance of low computing resource consumption and smooth control, we finally set a frame skip of k = 3 by relying on our experience.

\subsubsection{Prioritized replay}
In most cases, there was little replay memory with high rewards, which would be time-consuming with huge replay table and only sparse rewards. As prioritized replay was a method that can make learning from experience replay more efficient, we simply timed the important experiences in proportion to their rewards and stored them into the replay memory. That way, they would have a greater opportunity to be recalled \citep{schaul2015prioritized}.

\section{Experiments}
\label{sec:exp}
The model was trained using three different tracks, which cover all the track that Stanford used for comparison. An individual sets of weights was trained separately for each model because each track has different terrain textures. The rewards were low initially because it is equivalent to a random exploration at the beginning of training and because the driving agent would get stuck somewhere without making any significant progress. After about 1400 episodes of training, the driving agent finished the race. Under most circumstances, the driving agent did not finish the race in given steps so the reward was positive but not as high as receiving the final big reward. We set this step limit because of a lack of a reset mechanism for dead situations which was very useful in the early stage of training. In Stanford's report, they created a DNF flag to represent the driving agent getting stuck. In our experiment, the driving agent had learned better policy and displayed better behaviors, proving better robustness of the system. We also visualized the CNN in order to validate the ability of the model.

\subsection{Experimental Environment}

We chose the car racing game to carry out our simulated self-driving experiment. In order to play the car racing game autonomously, we used a third-partied OpenAI Gym environment wrapper for the game developed by Bzier\footnote{https://github.com/bzier/gym-mupen64plus}. The API provides direct access to the
game engine and allows us to run our scripts while playing the game frame-by-frame. Through the API, we can easily obtain the game information, either the screen output or the high-level intermediate game parameters like its location in the minimap. Our model was verified to be able to deal with the partially observable game environment efficiently.

\begin{figure*}[h!]
  \centering
  \includegraphics[width=1.1\linewidth]{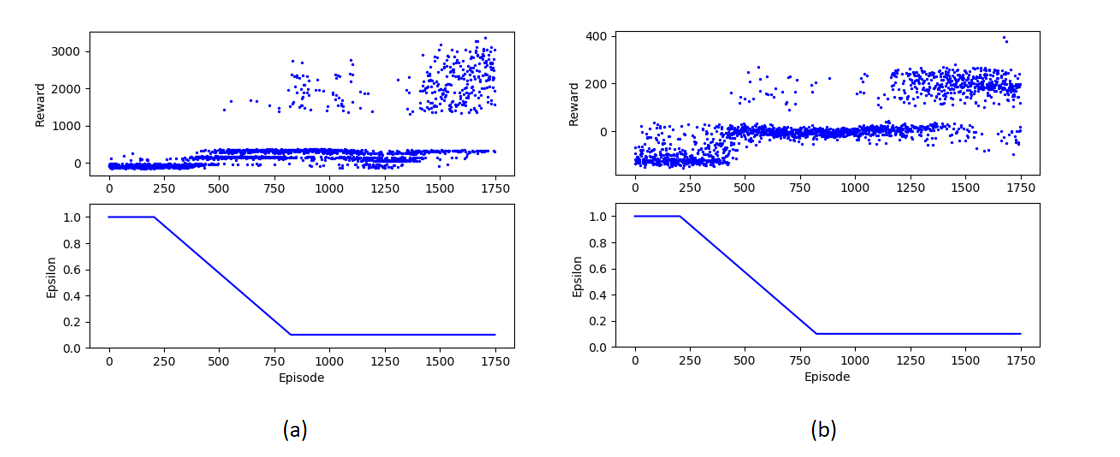}
  \caption{There are reward trends for two different maps. The picture (a) corresponded to Farm and the picture (b) corresponded to Raceway. After training for 1750 episodes, we obtained the reward tendency. At about 400 episodes, the stability of the driving agent began to increase. After 1400 episodes, the reward stabilized at a high level.   }
  \label{fig:5}
\end{figure*}

\subsection{Rewards Analysis}
For comparison, the model was trained and tested using the same tracks like those used by Stanford in their supervised learning method. Each track has different terrain textures and difficulty routes. Therefore, an individual set of weights was trained separately for each model. The experiments were carried out on a common configured portable laptop, and all models converged after spending over 80 hours each.

\subsection{CNN Visualization}
\begin{figure*}[h!]
  \centering
  \includegraphics[width=0.9\textwidth]{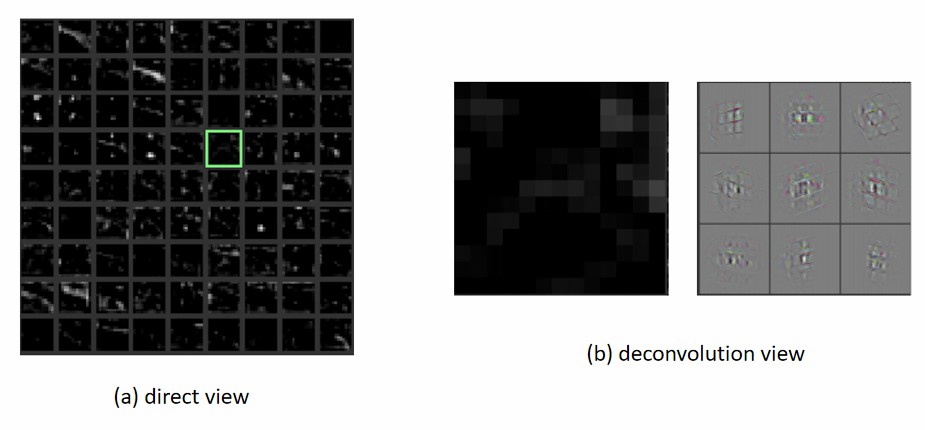}
  \caption{We visualized the last layer and obtained a direct view (left). Because the high-level layers output was quite abstract, we then visualized the last layer using deconvolution and obtained the right picture. It seemed to represent the wall element in the original graph.  }
  \label{fig:vislast}
\end{figure*}

The quality of the high-level feature was regarded as the key measurement of the learning process. The network could be evaluated from many aspects such as the loss function and the validation accuracy in traditional supervised learning. However, in unsupervised learning, we did not have this kind of standard to provide a quantitative assessment of the network. Hence, we visualized the output of the last layer (see Figure \ref{fig:vislast}), with the aim of ascertaining the kind of features that are captured by the network.The high-level layers output seemed quite abstract through direct observation. Thus, we visualized the high-level features through deconvolution. Deconvolution and unpooling are the common methods for reconstructing entity.

\subsection{Result and Discussion}

The test results are shown in Table \ref{tab:ler}. The driving agent was tested on several tracks in line with those of Stanford's experiment such as Farm, Raceway, and Mountain. We evaluated our model based on its achieved time. For each track, we run 10 races in real-time and calculate the mean race times as the final result. The Stanford results \citep{honeuralkart} and the Human results were borrowed from the Stanford report. The human results were obtained by a real human participant playing each track twice: The first time is to get used to the track, and the second time is to record their time. The results were obtained by testing the driving agent on each track ten times and then taking the average (excluding the DNF cases).

\begin{table*}[h!]
				\renewcommand{\arraystretch}{1.5}
				\footnotesize
				\caption{Performance comparison. }
				\setlength{\tabcolsep}{6mm}
				\centering
				\normalsize
				\begin{tabular}{l|c|c|c}
					\hline
					Track & Our model & Stanford model \citep{honeuralkart} & Human \\
				
					\hline
					Farm & 98.33 & 97.46 & 94.07 \\
					Raceway & 166 & 129.09, 1DNF & 125.30 \\
					Mountain  & 213 & 138.37, 2DNF & 129.50   \\ 
					\hline
				\end{tabular}
				\label{tab:ler} 
\end{table*}

In sum, our results seemed to require a longer period to finish the race. Through observation, we found that our agent’s driving path was not as smooth as that provided by supervised learning results. We closed this course for two reasons. First, we reduced the epsilon too fast. As shown in the Figure \ref{fig:5}, the epsilon value is rapidly adjusted to 0.1 as the rewards increase. This indicates the experiences show more reliability. Then during the latter phase of training, we depended mainly on experiences even though there were still many better state-action sets to explore. Second, we discretized the actions roughly. We used a set of [0; 20; 40; 80] as the options for speed. To be more precise, we used the control variable of the joystick. We observed that the numerical value of the joystick parameter did not maintain a linear relationship with the real effect on the driving agent as shown on the screen. Through observation, a speed of 20 m/s and 40 m/s both produced a tiny effect while a speed of 80m/s would make a radical change. An accurate modeling between the joystick parameter and the real effect would make the segmentation more reasonable. However, compared with the experiment done by the Stanford group, our experiments performed satisfactory even if the driving agent deviated from the driveway. We considered the brake  and trained the driving agent using a trial-and-error method, which was more in line with the real situation. Hence, the bio-inspired reinforcement learning method in the study was a more suitable approach for the driving agent to make decisions.

It is worth mentioning that the driving agent is more sensitive to the green grass than the yellow sand texture. The reason is that we used a fixed set of weights to transform the color image into a gray-scale image before inputting, which is in charge of this phenomenon. In the future, we will use the original three channels of the RGB data to carry out the experiments to end this conjecture and improve the results.

\section{Conclusion}
\label{sec:conclusion}
Brain-inspired learning has recently gained additional interests in solving decision-making tasks. In this paper, we propose an effective brain-inspired end-to-end learning method with the aim of controlling the simulated self-driving agent. Our modified DRQN model has proven to manage plenty of error states effectively, thus indicating that our trial-and-error method using deep recurrent reinforcement learning could achieve better performance and stability. By using the screen pixels as the only input of the system, our method highly resembles the experience of human-beings solving a navigation task from the first-person perspective. This resemblance makes this research inspirational for real-world robotics applications. Hopefully, the proposed brain-inspired learning system will inspire real-world self-driving control solutions.

\bibliography{references}{}
\bibliographystyle{IEEEtran}

\end{document}